\documentclass[sigconf,screen,nonacm]{acmart}

\usepackage{multirow}
\usepackage{arydshln}
\usepackage{colortbl}
\usepackage{subfigure}
\usepackage{float}
\usepackage{balance}
\usepackage[bottom]{footmisc}

\AtBeginDocument{%
  }

\acmSubmissionID{123-A56-BU3}
\acmSubmissionID{4874}

\begin{document}

\title{Benchmarking Retrieval-Augmented Generation in Multi-Modal Contexts}

\author{Zhenghao Liu}
\affiliation{%
  \institution{Northeastern University, China}
  \city{Shenyang}
  \country{China}
}
\email{liuzhenghao@mail.neu.edu.cn}

\author{Xingsheng Zhu}
\affiliation{%
  \institution{Northeastern University, China}
  \city{Shenyang}
  \country{China}
}
\email{zhuxingsheng@stumail.neu.edu.cn}

\author{Tianshuo Zhou}
\affiliation{%
  \institution{Northeastern University, China}
  \city{Shenyang}
  \country{China}
}
\email{zhoutianshuo.310@gmail.com}

\author{Xinyi Zhang}
\affiliation{%
  \institution{Northeastern University, China}
  \city{Shenyang}
  \country{China}
}
\email{delanyvv@163.com}

\author{Xiaoyuan Yi}
\affiliation{%
  \institution{Microsoft Research Asia}
  \city{Beijing}
  \country{China}
}
\email{xiaoyuanyi@microsoft.com}

\author{Yukun Yan}
\affiliation{%
  \institution{Tsinghua University}
  \city{Beijing}
  \country{China}
}
\email{yanyk.thu@gmail.com}
\authornote{Corresponding authors.}


\author{Ge Yu}
\affiliation{%
  \institution{Northeastern University, China}
  \city{Shenyang}
  \country{China}
}
\email{yuge@mail.neu.edu.cn}

\author{Maosong Sun}
\affiliation{%
  \institution{Tsinghua University}
  \city{Beijing}
  \country{China}
}
\email{sms@tsinghua.edu.cn}
\authornotemark[1]

\renewcommand{\shortauthors}{Zhenghao Liu et al.}

\settopmatter{printacmref=false}
\renewcommand\footnotetextcopyrightpermission[1]{} 
\pagestyle{plain}

\begin{abstract}
With the rapid advancement of Multi-modal Large Language Models (MLLMs), their capability in understanding both images and text has greatly improved. However, their potential for leveraging multi-modal contextual information in Retrieval-Augmented Generation (RAG) remains largely underexplored.
To address this gap, this paper introduces \textbf{M}ulti-\textbf{M}odal \textbf{R}etrieval-\textbf{A}ugmented \textbf{G}eneration (M$^2$RAG), a benchmark designed to evaluate the effectiveness of Multi-modal Large Language Models in leveraging knowledge from multi-modal retrieval documents. The benchmark comprises four tasks: image captioning, multi-modal question answering, multi-modal fact verification, and image reranking. All tasks are set in an open-domain setting, requiring RAG models to retrieve query-relevant information from a multi-modal document collection and use it as contextual input for RAG modeling. To enhance the context utilization capabilities of MLLMs, we also introduce \textbf{M}ulti-\textbf{M}odal \textbf{R}etrieval-\textbf{A}ugmented \textbf{I}nstruction \textbf{T}uning (MM-RAIT), an instruction tuning method that optimizes MLLMs within multi-modal contexts. Our experiments demonstrate the effectiveness of MM-RAIT by significantly improving the quality of responses generated by different RAG models, outperforming MiniCPM-V 2.6 and Qwen2-VL with 34\% and 33\% gains, respectively. All data and code are available at \url{https://github.com/NEUIR/M2RAG}.
\end{abstract}

\maketitle

\section{Introduction}
\begin{figure}[t]
    \centering
    \includegraphics[width=1\linewidth]{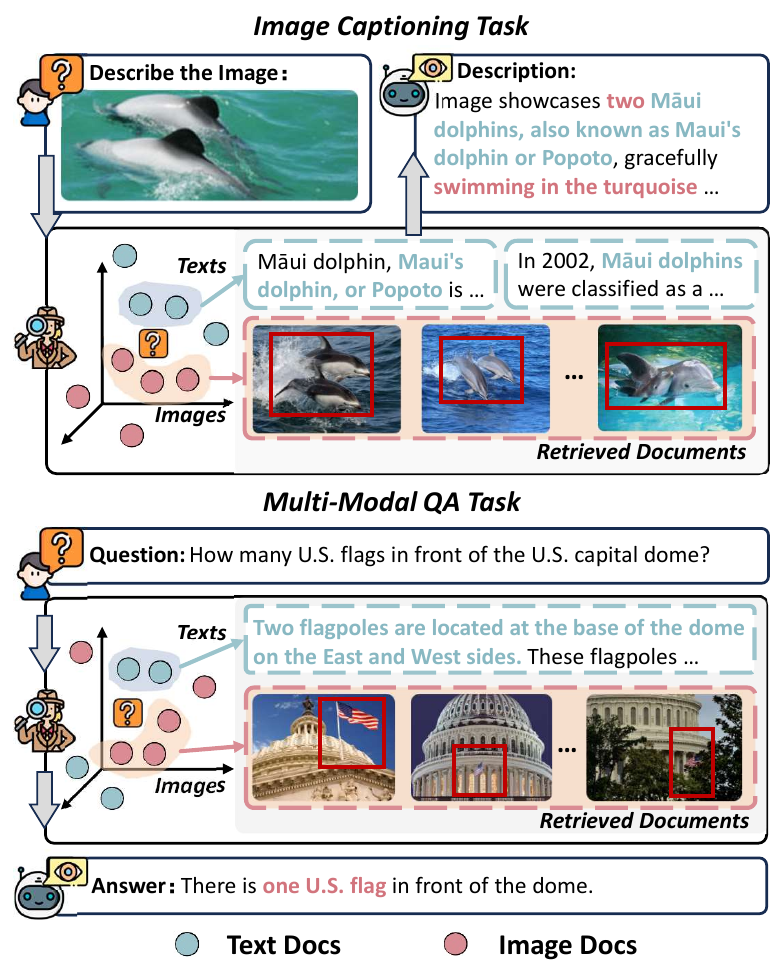}
    \caption{Illustration of Multi-Modal RAG Tasks. The documents, retrieved from different multi-modal knowledge sources, are used as contextual input to MLLMs.}
    \label{fig:mmrag}
\end{figure}
With the rapid development of Large Language Models (LLMs), such as GPT-4~\cite{openai2023gpt} and LLaMA~\cite{touvron2023llama} have demonstrated strong emergent abilities in many natural language processing (NLP) tasks~\cite{wei2022emergent, zhao2023survey}. However, LLMs often face the issue of hallucinations, causing them to produce unreliable responses~\cite{ji2023survey, huang2023survey, shuster2021retrieval}. Retrieval-Augmented Generation (RAG)~\cite{lewis2020retrieval, asai2024reliable, shi2023replug, yao2022react} has proven effective in mitigating this hallucination problem by incorporating external knowledge into the generation process, thereby improving the factual accuracy and reliability of LLM outputs.

To enhance LLMs with retrieved knowledge, existing approaches typically feed retrieved documents into LLMs as input contexts, prompting them to generate responses based on this in-context information~\cite{ram2023context}. Existing RAG approaches~\cite{petroni2021kilt, linra} usually focus on retrieving textual knowledge from corpora to aid LLMs in answering queries. Recent studies~\cite{hu2024mrag, sharifymoghaddam2024unirag, chen2022murag, caffagni2024wiki, ding2024ra, cui2024more} have extended Retrieval-Augmented Generation (RAG) to Multi-modal Large Language Models (MLLMs), enabling them to handle knowledge-intensive and information-seeking tasks involving visual queries. However, most existing benchmarks route queries to different models in an oracle manner, and evaluate MLLMs using either text or images as the sole external knowledge source. In contrast, real-world RAG scenarios often require retrieving query-relevant information from sources of diverse modalities~\cite{liu2022universal}, and effectively integrating complementary signals across modalities to generate accurate answers.


To advance RAG modeling in multi-modal scenarios, we introduce the Multi-Modal RAG (M$^2$RAG) benchmark, designed to explore the effectiveness of MLLMs by feeding multi-modal retrieved documents as the input contexts to answer the question. As shown in Figure~\ref{fig:mmrag}, we can first use images or text as queries to retrieve multi-modal documents via multi-modal dense retrievers~\cite{liu2022universal, zhou2024marvel, zhou2024vista}. Then these multi-modal documents are used as the input contexts to assist MLLMs to generate responses for the user query. Our benchmark emphasizes the integration of visual and textual modalities during both retrieval and generation stages. Different from existing works~\cite{aghajanyan2022cm3, sharifymoghaddam2024unirag}, M$^2$RAG is constructed based on high-quality datasets~\cite{chang2022webqa, mishra2022factify} and introduces four diverse tasks, image captioning, multi-modal question answering, multi-modal fact verification, and image reranking, for evaluation. To better reflect real-world scenarios, these tasks are reformulated in an open-domain setting, enabling a more comprehensive assessment of how effectively MLLMs can leverage knowledge from multi-modal contexts.

In this paper, we also propose the \textbf{M}ulti-\textbf{M}odal \textbf{R}etrieval \textbf{A}ug-mented \textbf{I}nstruction \textbf{T}uning (MM-RAIT) method to adapt MLLMs to the multi-modal in-context learning scenario, enhancing the effectiveness of MLLMs in utilizing the knowledge from these multi-modal retrieval documents. Specifically, we design task-specific prompt templates for different tasks in the M$^2$RAG benchmark and then fine-tune MLLMs within multi-modal retrieved context, making MLLMs maintain contextual awareness during generation. Our experimental results demonstrate that using retrieved knowledge significantly enhances MLLMs' performance, achieving significant improvements in both zero-shot and few-shot settings. After training with MM-RAIT, MiniCPM-V and Qwen2-VL show an average improvement of 34\% and 33\% over vanilla RAG modeling methods, showing the effectiveness of MM-RAIT.
\section{Related Work}
Existing RAG models~\cite{shi2023replug, asai2023self, yu2023augmentation, yan2024corrective} typically rely on dense retrievers~\cite{karpukhin2020dense, xiongapproximate, ren-etal-2021-rocketqav2, xiong2021answering, gao-callan-2022-unsupervised} or sparse retrievers~\cite{robertson2009probabilistic} for text document retrieval. Recent works~\cite{liu2022universal, zhou2024marvel,zhou2024vista} usually focus on broadening the effectiveness of text retriever to multi-modal retrieval scenarios, allowing the inclusion of rich external knowledge from different modalities within RAG frameworks. They build unified multi-modal retrieval systems that map images and texts into a shared semantic space, which allows for single-modal matching, cross-modal matching, and modality routing~\cite{liu2022universal}. These advancements enable the retrieval of multi-modal knowledge, providing a way for evaluating the effectiveness of MLLMs within multi-modal contexts.

Multi-modal Large Language Models (MLLMs)~\cite{achiam2023gpt,team2023gemini,sun2023emu,sun2024generative, aghajanyan2022cm3, lu2024deepseekvl} have proven their effectiveness in understanding, integrating, and utilizing both visual and textual knowledge in generation tasks. Models like BLIP~\cite{li2022blipbootstrappinglanguageimagepretraining, li2023blip}, LLaVA~\cite{liu2024visual}, and Flamingo~\cite{alayrac2022flamingo} build MLLMs by combining pre-trained vision encoders with Large Language Models (LLMs), enabling LLMs to process multi-modal inputs during generation. Emu2~\cite{sun2024generative} further extends the generative potential of MLLMs by pretraining using a large-scale multi-modal corpus with a unified autoregressive objective, thereby improving the model's transferability to a wide range of downstream tasks. Qwen-VL~\cite{Qwen2VL} further enhances multi-modal understanding and generation by integrating a high-resolution visual encoder with a fine-grained, multi-stage fusion mechanism, effectively aligning visual tokens with linguistic representations. In parallel, MiniCPM-V~\cite{yao2024minicpm} adopts a lightweight vision encoder and a compact LLM, offering a favorable trade-off between performance and efficiency, particularly suited for deployment in resource-constrained environments. Thriving on the advancements in MLLMs, researchers pay more attention to extending the advantages of Retrieval-Augmented Generation (RAG) to these MLLMs, enhancing their generation capability using the knowledge from different modalities. 

\begin{figure*}[t]
    \centering
    \includegraphics[width=1\linewidth]{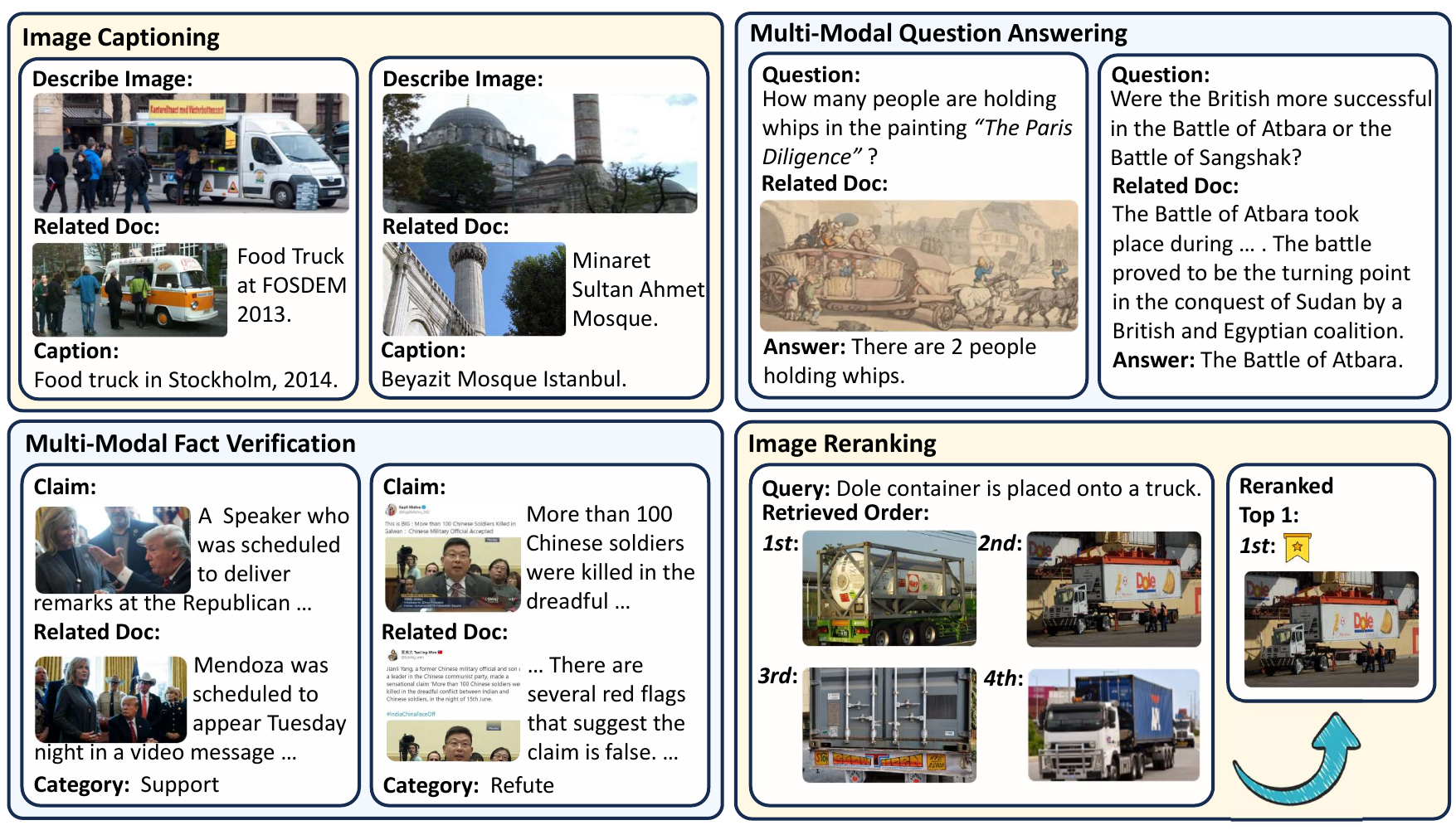}
    \caption{Examples of Different Tasks Defined in the M$^2$RAG Benchmark. All tasks are designed for the open-domain setting. Thus, we present the input, ground truth answers, and retrieved documents for each task.}
    \label{fig:case}
\end{figure*} 
Multi-modal RAG has demonstrated its potential to enhance knowledge-intensive and information-seeking tasks, such as question answering~\cite{chang2022webqa, marino2019ok} and fact verification~\cite{mishra2022factify}. These works~\cite{chen2022murag, caffagni2024wiki, ding2024ra, yu2024visrag} utilize retrieval-based multi-modal documents to provide richer and contextually relevant information. Wiki-LLaVA~\cite{caffagni2024wiki} enhances performance on Visual Question Answering (VQA) tasks by retrieving external knowledge based on the input image. 
VisRAG~\cite{yu2024visrag} further extends to document-level VQA tasks by directly leveraging document page images instead of extracted text, preserving and utilizing the original data within documents to enhance MLLM generation. RA-BLIP~\cite{ding2024ra} introduces an Adaptive Selective Knowledge Generation (ASKG) strategy, which enables the generator to autonomously assess the relevance of retrieved knowledge, thereby achieving strong denoising performance and effectively reducing the interference of irrelevant information during retrieval and generation. Additionally, several studies have applied multi-modal RAG to improve the performance of MLLMs on tasks like image captioning~\cite{lin2014microsoft, young2014image, hu2023reveal} and generation~\cite{yasunaga2023retrieval, yu2023scaling, sharifymoghaddam2024unirag}. MORE~\cite{cui2024more} further extends multi-modal RAG to commonsense reasoning tasks. During training, it introduces a ``query dropout'' method to prevent the language model from either completely ignoring the retrieved multi-modal documents due to potential noise or becoming overly reliant on possibly noisy retrieval results. While recent works have made significant progress in applying multi-modal RAG to various tasks, the evaluation of these systems remains underexplored. Existing multi-modal benchmarks~\cite{johnson2017clevr, schuhmann2021laion, lin2014microsoft, young2014image, marino2019ok} are typically tailored to specific tasks and lack a unified framework for systematically assessing the performance of multi-modal RAG models.

\section{M$^2$RAG Benchmark for Multi-Modal Retrieval-Augmented Generation} In this section, we introduce our Multi-Modal Retrieval-Augmented Generation (M$^2$RAG) benchmark. We first introduce the RAG tasks included in M$^2$RAG, followed by a detailed explanation of the construction process. Finally, we provide a comparative analysis between existing multi-modal benchmarks and M$^2$RAG.

\textbf{Task Definition.}\label{bench:task} As shown in Figure~\ref{fig:case}, M$^2$RAG defines four tasks to evaluate the capabilities of MLLMs in open-domain RAG scenarios: image captioning, multi-modal question answering, multi-modal fact verification, and image reranking. For each task, MLLMs are required to retrieve knowledge from the multi-modal document collection $\mathcal{D}$ and generate responses to answer the question $q$. 

\textit{Image Captioning Task.} Image Captioning is a widely used task for evaluating the performance of multi-modal RAG models~\cite{aghajanyan2022cm3, sharifymoghaddam2024unirag}. In this task, an image is provided as the query $q$, and the document collection $\mathcal{D}$ is constructed using image documents that contain captions. The goal of image captioning is to generate concise and semantically coherent captions that accurately describe the image content. Unlike previous works~\cite{aghajanyan2022cm3, sharifymoghaddam2024unirag}, we collect image captions from WebQA~\cite{chang2022webqa}, where all image documents are collected from Wikimedia Commons. These captions often include important details for verbalizing the semantics of images, such as named entities, making the task more challenging~\cite{liu2022universal}. 

\textit{Multi-Modal Question Answering Task.} Multi-Modal Question Answering (QA) is a task for assessing the capabilities of multi-modal RAG models in understanding and reasoning across both textual and visual modalities~\cite{chang2022webqa}. Given a textual query $q$, the model aims to generate accurate and informative answers by retrieving and leveraging relevant documents from a multi-modal collection $\mathcal{D}$, which includes text and image documents with captions.
We follow WebQA benchmark~\cite{chang2022webqa} and extend it to an open-domain setting, where the retriever selects query-relevant documents from the entire collection $\mathcal{D}$, following~\citet{liu2022universal}.

\textit{Multi-Modal Fact Verification Task.} The Multi-Modal Fact Verification task challenges MLLMs to verify the accuracy of claims using retrieved multi-modal evidence. In this task, the query $q$ can be a multi-modal claim, and the document collection $\mathcal{D}$ consists of both text and image documents, where the image documents do not contain captions. Each claim is assigned one of three labels, ``Support'', ``Refute'', or ``Insufficient'', indicating whether the retrieved evidence supports, refutes or lacks sufficient information to verify the claim. We build this task on the Factify dataset~\cite{mishra2022factify}, but we focus on open-domain fact verification by retrieving evidence from a multi-modal document collection~\cite{thorne2018fever}.

\textit{Image Reranking Task.} In the Image Reranking task, the objective is to identify the most relevant images based on a given image description. In this setting, the image description serves as query $q$, and the document collection $\mathcal{D}$ consists of image documents without associated captions. For each description, we first use a multi-modal retriever to retrieve candidate image documents based solely on their image features and then rerank the images using MLLMs. To adapt MLLMs for this task, we follow previous work~\cite{muennighoff2022sgpt} and compute the Perplexity (PPL) score to rerank image candidates based on their image features. This approach models the relevance between queries and images in a manner similar to image captioning, where a lower PPL score indicates greater relevance between the candidate image and the given query.

\begin{table}[t]
\centering
    \caption{Comparison of Multi-Modal Benchmarks.}
    \label{tab:benchmarks_comparison}
      \resizebox{\linewidth}{!}{
    \begin{tabular}{lccccc}
    \hline
    \textbf{Benchmarks} & \textbf{Input} & \textbf{Retrieval} & \textbf{Multi Task} & \textbf{Open Domain}\\
    \hline
    MSCOCO~\shortcite{lin2014microsoft} & Image   & \includegraphics[height=0.8em]{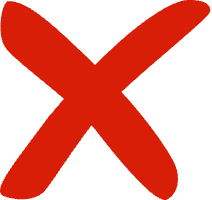} & \includegraphics[height=0.8em]{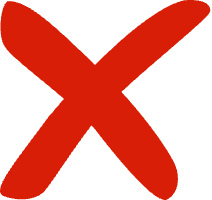}  & \includegraphics[height=0.8em]{samples/latex/image/error.png} \\
    Flickr30K~\shortcite{young2014image} & Image  & \includegraphics[height=0.8em]{samples/latex/image/error.png} & \includegraphics[height=0.8em]{samples/latex/image/error.png}  & \includegraphics[height=0.8em]{samples/latex/image/error.png}\\
    K-VQA~\shortcite{shah2019kvqa} & Multi  & \includegraphics[height=0.8em]{samples/latex/image/error.png} & \includegraphics[height=0.8em]{samples/latex/image/error.png}  & \includegraphics[height=0.8em]{samples/latex/image/error.png}\\
    WebQA~\shortcite{chang2022webqa}& Text  & Multi & \includegraphics[height=0.8em]{samples/latex/image/error.png} & \includegraphics[height=0.8em]{samples/latex/image/error.png} \\
    MRAG-Bench~\shortcite{hu2024mrag} & Multi &  Image& \includegraphics[height=0.8em]{samples/latex/image/error.png}  & \includegraphics[height=0.8em]{samples/latex/image/error.png}\\ \hline
    \textbf{M$^2$RAG (Ours)} & Multi & Multi & \includegraphics[height=0.8em]{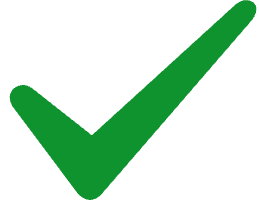} & \includegraphics[height=0.8em]{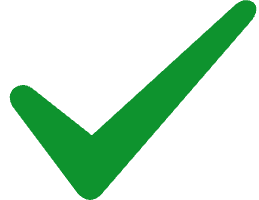}\\
    \hline
    \end{tabular}}
\end{table}
\textbf{Details of Data Construction.} To build the M$^2$RAG benchmark, we collect data from two datasets, WebQA~\cite{chang2022webqa} and Factify~\cite{mishra2022factify}. 

We adapt the WebQA dataset to construct task-specific benchmarks for image captioning, multi-modal question answering, and image reranking. For multi-modal QA, we sample equal numbers of text- and image-based QA pairs to ensure modality balance. For image captioning and reranking, we randomly select image-text pairs with similarity >0.65, splitting them into training and test sets, with the same image-caption pairs used in both test sets.
The retrieval corpus for multi-modal QA follows the original WebQA setup, and the image reranking task shares this image corpus. To prevent data leakage, images used in training or evaluation are excluded when constructing the captioning retrieval corpus.


For the multi-modal fact verification task, since the test labels in Factify~\cite{mishra2022factify} are unavailable, we follow~\citet{tahmasebi2024multimodal} and sample from the validation set to construct the evaluation set. All text and image documents from Factify’s training and validation sets are collected to build the retrieval corpus. The original Factify dataset consists of five categories: ``Support\_Text'', ``Support\_Multimodal'', ``Insufficient\_Text'', ``Insufficient\_Multimodal'', and ``Refute''. When constructing the training and evaluation datasets for M$^2$RAG, we select an equal number of samples from each of these five categories to maintain class balance. Since our RAG scenario involves both text and image information, we merge modality-specific labels into three unified classes: ``Support'', ``Refute'', and ``Insufficient''.

\begin{table}[t]
\centering
\caption{\label{tab:img_cap_compare}Performance of Different MLLMs in the Image Captioning Tasks of MSCOCO and M$^2$RAG.}
\begin{tabular}{l|cccc}
\hline
\textbf{Benchmark} & \textbf{BLEU-2} & \textbf{BLEU-4} & \textbf{ROUGE-L} & \textbf{CIDEr} \\ \hline
\rowcolor{gray!8} \multicolumn{5}{l}{\textbf{MiniCPM-V 2.6 (8B)}} \\ \hline
MSCOCO & 11.53 & 4.19 & 30.68 & 33.48 \\
M$^2$RAG  & 4.62 & 1.91 & 17.58 & 18.39 \\ \hline
\rowcolor{gray!8} \multicolumn{5}{l}{\textbf{Qwen2-VL (7B)}} \\ \hline
MSCOCO & 16.93 & 6.75 & 37.51 & 70.75 \\
M$^2$RAG & 5.08 & 2.24 & 19.48 & 26.01 \\ \hline
\end{tabular}%

\end{table}

\begin{figure}[t]
    \centering
    \includegraphics[width=1\linewidth]{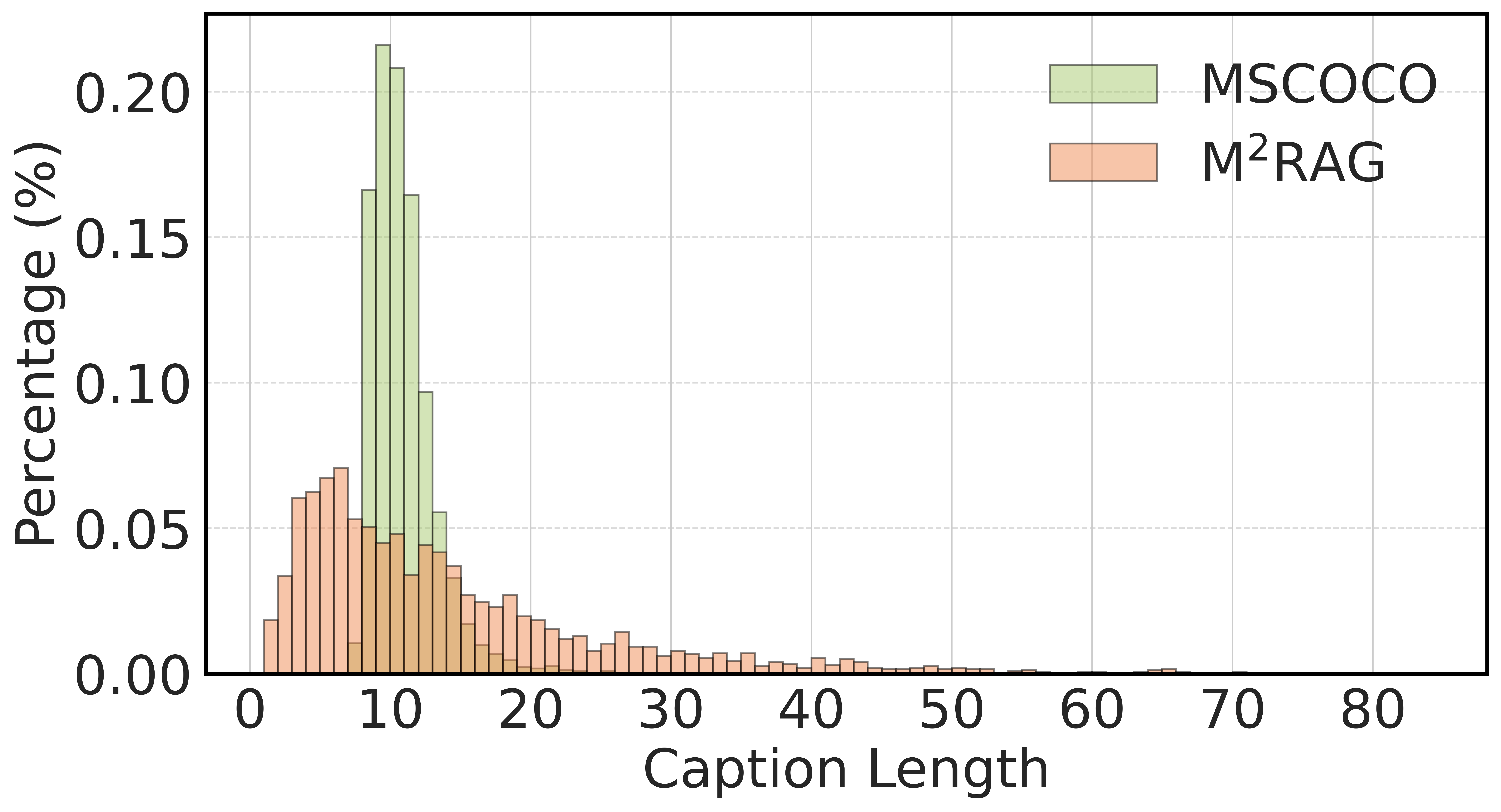}
    \caption{Length Distribution of Captions in the MSCOCO and M$^2$RAG Benchmarks.}
    \label{fig:img_cap_compare}
\end{figure}

\textbf{Benchmark Comparison.} The comparison between existing benchmarks and M$^2$RAG is presented in Table~\ref{tab:benchmarks_comparison}.

Most existing multi-modal benchmarks focus on single tasks like image captioning~\cite{lin2014microsoft, young2014image} or QA~\cite{chang2022webqa}, and typically lack a retrieval component, limiting their evaluation of MLLMs in multi-modal RAG scenarios. In contrast, M$^2$RAG provides: 1) M$^2$RAG defines four tasks that assess an MLLM's ability to effectively understand and utilize retrieved knowledge. These tasks require MLLMs to perform reasoning and information matching based on both queries and contextual knowledge. 2) M$^2$RAG incorporates the multi-modal retrieval results as the contexts for model input, avoiding the need for separate processing of the retrieval documents of different modalities. 3) M$^2$RAG adapts these tasks to an open-domain setting, offering a more realistic and challenging RAG scenario compared to existing benchmarks that rely on closed or narrow-domain data.


To further illustrate the difficulty posed by M$^2$RAG, we additionally compare the performance of MiniCPM-V 2.6 and Qwen2-VL on the image captioning task using the MSCOCO~\cite{lin2014microsoft} and M$^2$RAG datasets. For MSCOCO, we use the version employed in the image captioning task of UniRAG~\cite{sharifymoghaddam2024unirag} and follow the same processing method described in their paper. As shown in Table~\ref{tab:img_cap_compare}, both MiniCPM-V 2.6 and Qwen2-VL exhibit lower performance on M$^2$RAG compared to MSCOCO, with average declines of over 9\% and 19\%, respectively. This suggests image captioning task in M$^2$RAG is more challenging for MLLMs than in MSCOCO.
As illustrated in Figure~\ref{fig:img_cap_compare}, captions in M$^2$RAG are more diverse in length and content, vary with the complexity of the scene and contain richer descriptions of entities and detailed contextual information. Compared to the formulaic captions in MSCOCO, M$^2$RAG requires deeper semantics and contextual reasoning, pushing MLLMs to utilize external knowledge for accurate and context-aware captions. In contrast, MSCOCO captions are simpler and allow MLLMs to rely mainly on internal knowledge. This highlights M$^2$RAG’s value as a more challenging benchmark for multi-modal RAG.



\section{Instruction Tuning for Multi-Modal Retrieval-Augmented Generation}
In this section, we present our Multi-Modal Retrieval-Augmented Instruction Tuning (MM-RAIT) method. First, we describe the framework for multi-modal Retrieval-Augmented Generation (RAG) (Sec.~\ref{method:rag}). Then, we introduce the multi-task instruction tuning method to enhance the performance of MLLMs in multi-modal RAG tasks (Sec.~\ref{method:mmtuning}).

\subsection{The Framework of Multi-Modal Retrieval-Augmented Generation}
\label{method:rag}

Given a query $q$, multi-modal RAG models first employ a retriever to search for query-relevant multi-modal documents $\mathcal{D}$ and then feed these documents to MLLMs to assist them in answering the query $q$. Each document $d \in \mathcal{D}$ can be either an image document or a text document. The multi-modal RAG framework consists of two main components: the multi-modal retrieval module and the retrieval-augmented generation module.

\textbf{Multi-Modal Retrieval.}
To retrieve documents from the multi-modal document collection $\mathcal{D}$, existing methods typically rely on multi-modal dense retrieval models~\cite{zhou2024marvel,zhou2024vista}.

Given a query $q$ and a multi-modal document $d$, multi-modal dense retrieval models, such as VISTA~\cite{zhou2024vista}, encode both as representations $h_q$ and $h_d$, respectively, and map them into an embedding space for retrieval: 
\begin{equation}
h_q = \text{Enc} (q); h_d = \text{Enc}(d), 
\end{equation} 
where $\text{Enc}$ denotes the encoder model. The query $q$ can be either a text or an image, and the multi-modal document $d$ can be a text or an image document. For documents containing captions, both image features and image captions are fed into the encoder model.

Next, we compute the similarity score $S(q,d)$ between the representations $h_q$ and $h_d$ of the query and document:
\begin{equation}
S(q,d) = \text{Sim}(h_q,h_d), 
\end{equation} 
where $\text{Sim}$ denotes cosine similarity. We then perform a KNN search~\cite{johnson2019billion} to retrieve the top$k$ most relevant multi-modal documents $\Tilde{\mathcal{D}}=\{d_1,...,d_k\}$ to the query $q$. During retrieval, the multi-modal retriever needs to conduct single-modality matching, cross-modality matching and modality routing in the embedding space~\cite{liu2022universal}.

\textbf{Multi-Modal RAG Module.}
After retrieval, we input the retrieved documents $\Tilde{\mathcal{D}}$ and query $q$ into the MLLM ($\mathcal{M}$), such as MiniCPM-V~\cite{yao2024minicpm} or Qwen2-VL~\cite{Qwen2VL}, to generate the output $y$:
\begin{equation}\label{eq:rag_module}
y = \mathcal{M}(\Tilde{\mathcal{D}},q). 
\end{equation} 
These retrieved documents provide external knowledge, which helps to update the parametric memory of the MLLM, enabling it to generate more accurate responses to the query $q$.

\subsection{MM-RAIT: Multi-Task Multi-Modal Instruction Tuning for MLLMs}
\label{method:mmtuning}
To adapt MLLMs to the multi-modal RAG scenario, we propose the \textbf{M}ulti-\textbf{M}odal \textbf{R}etrieval-\textbf{A}ugmented \textbf{I}nstruction \textbf{T}uning (MM-RAIT) method, designed to further enhance the performance of MLLMs across various RAG tasks.

To improve the MLLM generation process, we incorporate external knowledge to assist in answering the query (Eq.~\ref{eq:rag_module}). Specifically, we follow previous work~\cite{ram2023context} and concatenate the representations of the retrieved documents $\Tilde{\mathcal{D}}$ along with the query $q$ as the input for the MLLM ($\mathcal{M}$) to generate the output $y$: 
\begin{equation}
y = \mathcal{M}(\text{Instruct}_p,X(\Tilde{\mathcal{D}}),q), 
\end{equation} 
where $\text{Instruct}_p$ is the instruction for the task $p$, and $X(\Tilde{\mathcal{D}})$ denotes the concatenation of the representations of the retrieved documents: 
\begin{equation}
X(\Tilde{\mathcal{D}}) = X(d_1) \oplus \dots \oplus X(d_k). 
\end{equation} 
For the $i$-th retrieved document $d_i$, its representation can be the text sequence for a text document, the image features for an image document, or the concatenation of both image features and caption for an image document that contains a caption.

Next, we gather queries from three tasks to form the query set $Q$: image captioning, multi-modal question answering, and multi-modal fact verification. For each query $q$ in these tasks, the training objective for the model is to minimize the negative log-likelihood of generating the target sequence $y^*$: 
\begin{equation}
\mathcal{L} = - \sum_{q \in Q} \sum_{t=1}^T \log P(y^*_{t} \mid y^*_{<t}, \Tilde{\mathcal{D}}, q; \theta),
\end{equation} 
where $T$ is the length of the ground truth response, $y^*_{t}$ denotes the $t$-th token of the ground truth response, and $\theta$ represents the parameters of the MLLM ($\mathcal{M}$).
\section{Experimental Methodology} This section outlines the datasets, evaluation metrics, baselines, and implementation details used in our experiments.

\begin{table}[t]
\centering
\small
\caption{\label{tab:data_statistic}Data Statistics of M$^2$RAG.}
\resizebox{\linewidth}{!}{%
\begin{tabular}{l c r r r r}\toprule
\multirow{2}{*}{\textbf{Task}} & \multirow{2}{*}{\textbf{Source}} & \multicolumn{2}{c}{\textbf{\#Docs}} & \multicolumn{2}{c}{\textbf{\#Query}} \\ \cline{3-6}
& &{Image} & {Text} &{Train} & {Test}\\ \hline
\textbf{Img. Captioning} & WebQA & 383,750 & - & 3,000 & 3,000\\
\textbf{MM QA} & WebQA & 389,750 & 544,459 & 3,000 & 3,000 \\
\textbf{MM FV} & Factify & 41,000 & 41,000 & 3,000 & 3,000\\
\textbf{Img. Reranking} & WebQA & 389,750 & - & - & 3,000 \\ \bottomrule
\end{tabular}}
\end{table}

\textbf{Dataset.}
We use the M$^2$RAG dataset to evaluate the performance of different MLLMs in the multi-modal RAG scenario.
Detailed data statistics are shown in Table~\ref{tab:data_statistic}. For multi-modal retrieval, we adopt VISTA~\cite{zhou2024vista}, a universal multi-modal embedding model designed to retrieve query-related documents, enabling flexible processing of both text and image data inputs.

\begin{table*}[t]
\centering
\caption{\label{tab:overall}Overall Performance. We evaluate the performance of different RAG models implemented with MiniCPM-V 2.6 and Qwen2-VL on our M$^2$RAG benchmark. For the Image Reranking task, topK indicates reranking the K most relevant retrieved images. For other tasks, topK denotes retrieving the K most relevant documents as input contexts.}
\resizebox{\textwidth}{!}{%
\begin{tabular}{l|ccc|ccc|cc|c}
\hline
\multirow{2}{*}{\textbf{Model}} & \multicolumn{3}{c|}{\textbf{Image Captioning}} & \multicolumn{3}{c|}{\textbf{Multi-Modal QA}} & \multicolumn{2}{c|}{\textbf{MM Fact Verification}} & \textbf{Image Reranking} \\ \cline{2-10} 
& \textbf{BERTScore} & \textbf{ROUGE-L}& \textbf{CIDEr}& \textbf{BERTScore} & \textbf{ROUGE-L}& \textbf{CIDEr}& \textbf{ ACC }& \textbf{F1}& \textbf{FID↓} \\ \hline

\textbf{Qwen2.5-VL (7B)} & 85.89 & 19.94 & 22.07 & 78.18 & 37.69	& 139.96 & 47.70 & 39.64 & - \\ \cdashline{1-10}
w/ top5 & 86.73 & 23.91 & 44.16 & 77.91 & 42.28 & 178.97 & 54.03 & 48.64 & 10.29 \\ \hline

\textbf{InternVL 3 (8B)}& 85.45 & 17.15	& 18.87 & 78.02	& 35.51 & 106.09 & 45.47 & 45.08 & - \\ \cdashline{1-10}
 w/ top5 & 87.23	& 28.52	& 65.65 & 78.36	& 43.26	& 182.04 & 59.67 & 59.86 & 11.20 \\ \hline

\textbf{LLaVA-NeXT (7B)} & 86.55	& 17.04 & 16.59 & 78.68 & 47.33	& 254.82 & 42.00 & 34.76 & - \\ \cdashline{1-10}
w/ top5 & 87.48 & 24.03 & 37.14 & 78.69 & 49.82 & 263.71 & 53.60 & 44.35 & 10.51 \\ \hline

\textbf{GPT-4o-mini} & 85.24	& 17.11 & 15.36 & 77.92 & 33.92	& 85.37 & 53.67 & 49.20 & - \\ \cdashline{1-10}
w/ top5 & 84.91 & 17.24 & 16.85 & 78.06 & 39.10 & 116.02 & 61.33 & 59.31 & - \\  \hline

\textbf{MiniCPM-V 2.6 (8B)} & 85.55	& 17.58 & 18.39 & 77.89 & 32.78	& 82.65 & 43.03 & 41.13 & - \\ \cdashline{1-10}
w/ top1 & 86.80 & 24.28 & 43.76 & 78.09	& 37.87	& 120.13 & 54.83 & 53.49 & 12.17 \\ 
w/ top3& 86.67 & 23.13	& 37.89 & 78.03 & 38.34 & 121.05 & 56.33 & 54.01 & 10.86 \\ 
w/ top5 & 86.62 & 22.82 & 36.09 & 77.97 & 37.97 & 114.56 & 55.33 & 52.69 & 10.95 \\ \cdashline{1-10} 
\textbf{MM-RAIT} w/ top5 & \textbf{88.50} & \textbf{35.61} & \textbf{89.18} & \textbf{80.52} & \textbf{61.32} & \textbf{318.39} & \textbf{61.60} & \textbf{58.47} & \textbf{9.28} \\ \hline

\textbf{Qwen2-VL (7B)}& 86.32 & 19.48	& 26.01 & 78.32	& 39.05 & 153.40 & 45.43 & 34.23 & - \\ \cdashline{1-10}
w/ top1& 87.34	& 25.43	& 46.32 & 78.27 & 42.13 & 187.98 & 51.60 & 41.05 & 12.17 \\ 
w/ top3& 87.40 & 25.70	& 45.08 & 78.18	& 42.16 & 179.83 & 52.43 & 41.94 & 9.89 \\ 
w/ top5& 87.29	& 25.31	& 44.45 & 78.17	& 42.38	& 179.99 & 52.00 & 41.64 & 9.81 \\ \cdashline{1-10}
\textbf{MM-RAIT} w/ top5 & \textbf{89.12}	& \textbf{39.18}	& \textbf{118.00} & \textbf{80.46} & \textbf{63.45} & \textbf{339.13} & \textbf{62.13}	& \textbf{55.90} & \textbf{8.55} \\ \hline
\end{tabular}%
}

\end{table*}

\textbf{Evaluation Metrics.}
For image captioning and multi-modal QA tasks, we use BERTScore~\cite{zhangbertscore}, CIDEr~\cite{vedantam2015cider} and ROUGE~\cite{lin2004rouge} scores to assess performance. In the multi-modal fact verification task, we evaluate the performance of different RAG models using accuracy (ACC) and F1 score. For the image reranking task, we use the Fréchet Inception Distance (FID↓)~\cite{heusel2017gans}\footnote{\url{https://github.com/mseitzer/pytorch-fid}} for evaluation. 

\textbf{Baselines.}
We compare our models with various open-source multi-modal baselines, including Qwen2.5-VL~\cite{bai2025qwen2}, InternVL 3~\cite{zhu2025internvl3} and LLaVA-NeXT~\cite{li2024llava}, as well as our primary baselines MiniCPM-V 2.6~\cite{yao2024minicpm} and Qwen2-VL~\cite{Qwen2VL}. Additionally, we evaluate the API-based model GPT-4o-mini~\cite{hurst2024gpt} for reference. We apply MM-RAIT to MiniCPM-V 2.6 and Qwen2-VL for fine-tuning within the RAG framework and evaluate their performance by incorporating the top1, top3, and top5 retrieved documents as input. For the other baselines, we evaluate the performance of both the vanilla model and the RAG-enhanced model with top5 documents. For GPT-4o-mini, we randomly select 300 instances per task for evaluation.

\textbf{Implementation Details.}
We employ the Low-Rank Adaptation (LoRA)~\cite{hu2021lora} method and use LLaMA-Factory~\cite{zheng2024llamafactory} to fine-tune both MiniCPM-V 2.6 and Qwen2-VL using the top5 retrieved multi-modal documents for 2 epochs. The batch size is 4, with a maximum token limit of 4,096. A cosine learning rate scheduler is used, with the learning rate set to $5e-5$ for MiniCPM-V and $1e-4$ for Qwen2-VL. We set the \texttt{max\_pixels} parameter of Qwen2-VL to $512 \times 512$ during training and inference.
\begin{table*}[t]
\small
\centering
\caption{\label{tab:ablation}Ablation Study. We evaluate the performance of different retrieval modalities for candidate corpora on M$^2$RAG benchmark. For Image Captioning and Multi-Modal QA, we use ROUGE-L as the evaluation metric and F1-score is used for the MM Fact Verification task.}
\resizebox{\textwidth}{!}{%
\begin{tabular}{l|c|ccc|ccc|ccc}
\hline
\multirow{2}{*}{\textbf{Model}} & \multirow{2}{*}{\textbf{\#Doc}} & \multicolumn{3}{c|}{\textbf{Image Captioning}} & \multicolumn{3}{c|}{\textbf{Multi-Modal QA}} & \multicolumn{3}{c}{\textbf{MM Fact Verification}} \\ \cline{3-11} 
& &\textbf{Only Text} & \textbf{Only Image}& \textbf{Multi} & \textbf{Only Text} & \textbf{Only Image}& \textbf{Multi}& \textbf{Only Text} & \textbf{Only Image}& \textbf{Multi} \\ \hline
\textbf{MiniCPM-V 2.6 (8B)}& 0 & 17.58 & 17.58& 17.58 & 32.78 & 32.78	& 32.78 & 41.13 & 41.13 & 41.13 \\ \cdashline{1-11}
\multirow{3}{*}{\textbf{MM-RAIT}}
&1 & 33.17 & 32.21     & 34.58     & 57.75     & 52.29      & 58.54     & 59.11     & 51.52     & 58.55     \\
&3 &34.78& 31.96 & \textbf{35.68} & \textbf{60.35} & 51.94 & 61.05 & 59.72 & \textbf{52.55} & \textbf{59.27} \\ 
& 5 & \textbf{35.20} & \textbf{32.28} & 35.61 & 60.03 & \textbf{52.88} & \textbf{61.32} & \textbf{60.19} & 52.43 & 58.47 \\ \hline

\textbf{Qwen2-VL (7B)} & 0 & 19.48 & 19.48 & 19.48 & 39.05 & 39.05	& 39.05 & 34.23 & 34.23 & 34.23 \\ \cdashline{1-11}
\multirow{3}{*}{\textbf{MM-RAIT}} 
&1 & 37.19     & 35.63     & 37.95     & 60.36     & 53.43      & 60.62     & 52.55     & 45.48     & 53.06     \\
&3 & 38.00 & 35.65 & 39.05 & 62.45 & 53.92	& 63.02 & \textbf{52.71} & 47.15 & 55.57 \\ 
&5 & \textbf{38.37} & \textbf{35.77} & \textbf{39.18} & \textbf{62.80} & \textbf{53.96} & \textbf{63.45}	& 52.63 & \textbf{47.98} & \textbf{55.90} \\ \hline
\end{tabular}%
}

\end{table*}

\section{Evaluation Result}
In this section, we first evaluate MLLMs on the M$^2$RAG benchmark and conduct ablation studies on the impact of varying the number of retrieved documents across modalities. We then analyze the role of each retrieval modality in RAG and conclude with case studies.

\subsection{Overall Performance}
As shown in Table~\ref{tab:overall}, we report the performance of various RAG models on the M$^2$RAG benchmark. The zero-shot setting generates the output based on the vanilla MLLM only, and vanilla RAG models directly use retrieved documents to augment MLLMs, while MM-RAIT models fine-tune MLLMs within the RAG framework.

For these vanilla RAG models, performance generally improves as the number of retrieved documents increases. However, when retrieving the top5 ranked documents, the overall performance of vanilla RAG models on most tasks is lower compared to using the top1 or top3 documents. This highlights their difficulty in effectively filtering and integrating multi-modal information. Although some related works also use image captioning tasks to evaluate RAG performance~\cite{sharifymoghaddam2024unirag}, the performance of these MLLMs on M$^2$RAG is considerably worse, indicating that M$^2$RAG offers a more challenging dataset for image captioning.
Unlike vanilla RAG models, MiniCPM-V 2.6 and Qwen2-VL show strong performance on M$^2$RAG after MM-RAIT training. Qwen2-VL achieves over a 33\% average improvement, while MiniCPM-V 2.6 reaches 34\%, demonstrating MM-RAIT’s effectiveness in enhancing multi-modal context utilization for generation.

\begin{figure}[t]
    \centering
    \subfigure[MiniCPM Performance on Text Answerable Queries.\label{fig:text_qa_cpmv}]{
        \includegraphics[width=0.48\linewidth]{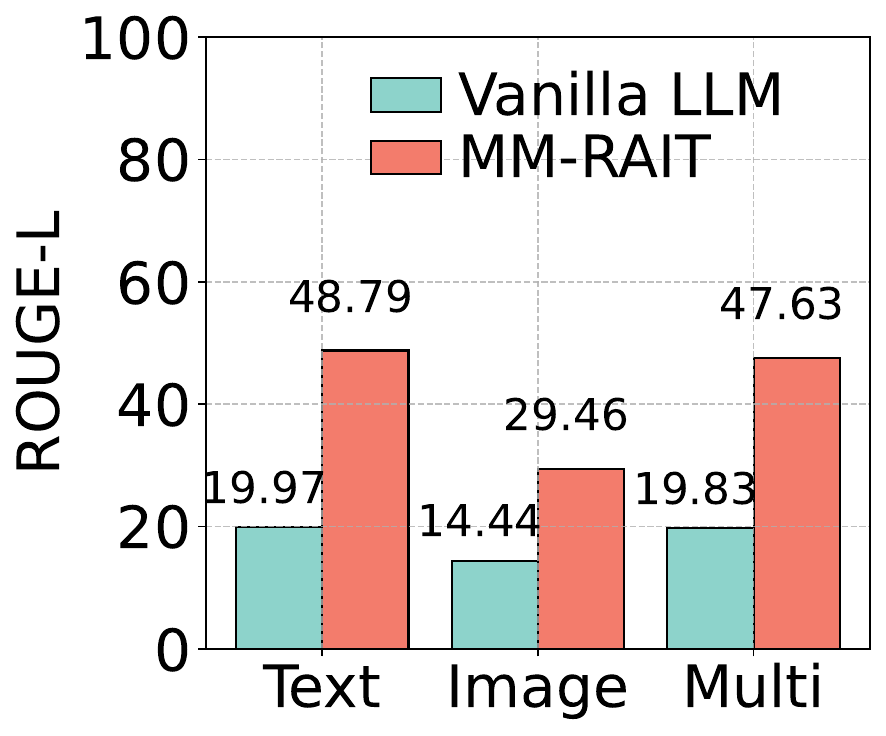}}
    \subfigure[Qwen2 Performance on Text Answerable Queries.\label{fig:text_qa_qwen}]{
        \includegraphics[width=0.48\linewidth]{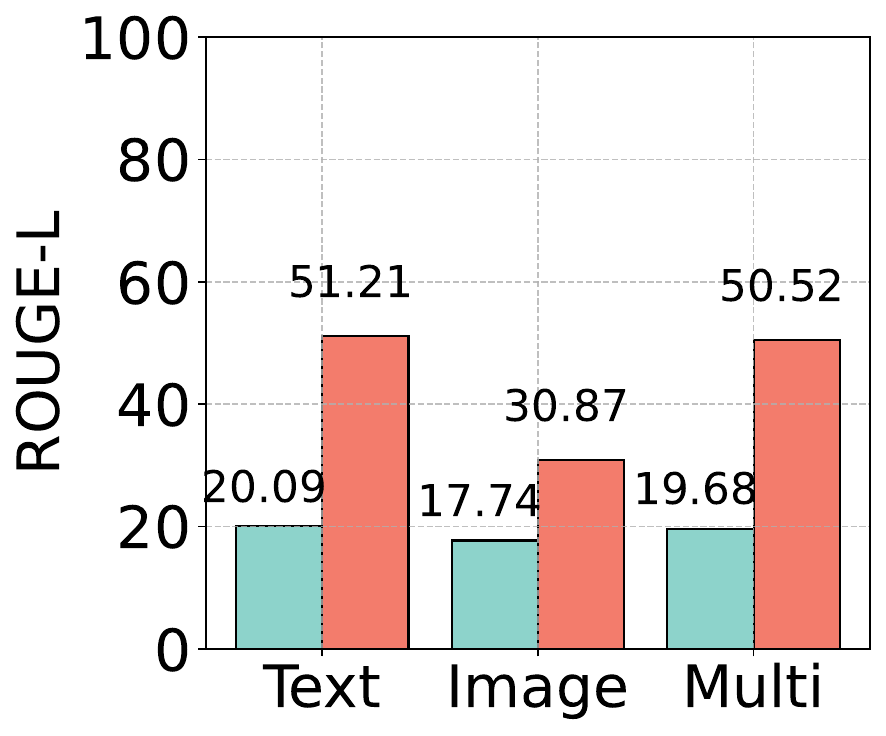}}
    \subfigure[MiniCPM Performance on Image Answerable Queries.\label{fig:img_qa_cpmv}]{
        \includegraphics[width=0.48\linewidth]{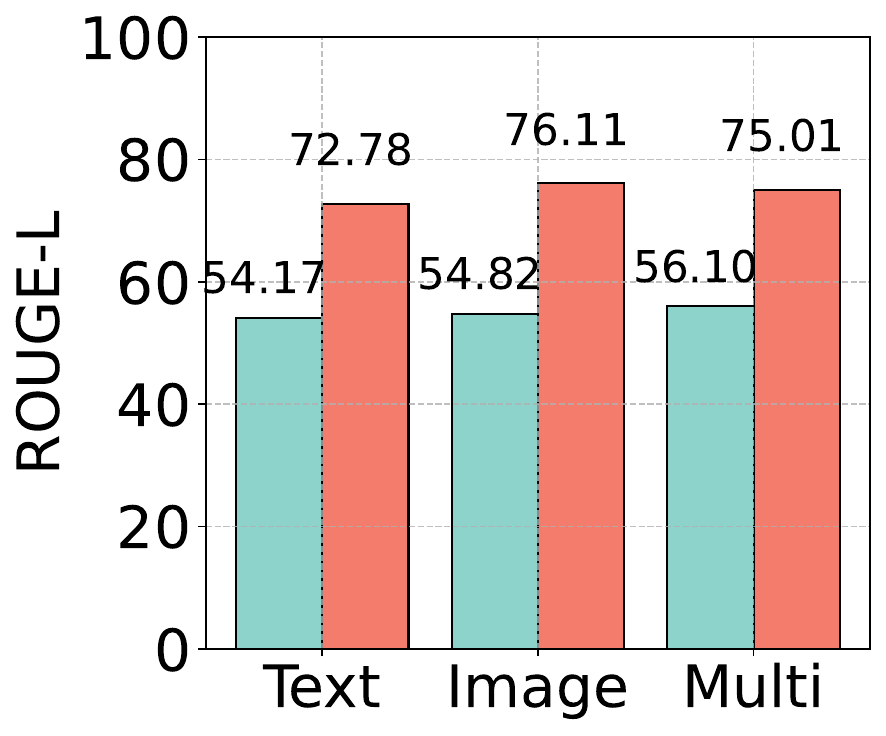}}
    \subfigure[Qwen2 Performance on Image Answerable Queries.\label{fig:img_qa_qwen}]{
        \includegraphics[width=0.48\linewidth]{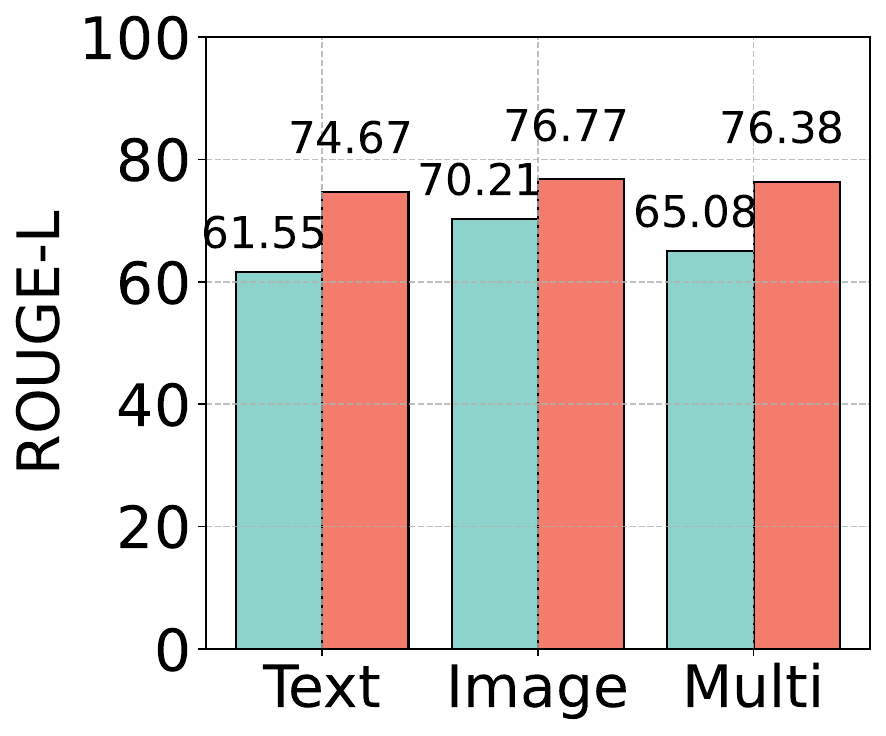}}
    \caption{RAG Performance in Multi-Modal QA Task Using Retrieved Documents of Different Modalities. 
    Text, Image, and Multi denote that retrieved text, image, and multi-modal documents are fed to different RAG models for evaluation.}
    \label{fig:retrieval_modality_comparison}
\end{figure}
\subsection{Ablation Study}
As shown in Table~\ref{tab:ablation}, we perform ablation studies to assess RAG effectiveness across different modalities and document counts. 

Specifically, we evaluate two settings: Only Text, which removes image features, and Only Image, which removes text from top-ranked multi-modal inputs, to isolate the contribution of each modality.
Compared with the RAG models using top3 ranked multi-modal documents for augmentation, the performance of vanilla RAG models usually decreases with top5 ranked documents, while MM-RAIT alleviates the performance decrease but also shows limited improvements. It illustrates that effectively using the multi-modal context is still challenging. Moreover, we further remove all texts or image features to show the roles of different modalities in RAG modeling. For all tasks, the RAG performance of the Only Text model slightly decreases, indicating text is the primary knowledge source for MLLMs. After adding the image features, the RAG performance usually increases, showing that image features can improve the performance of RAG models. Even though different modalities show the effectiveness in multi-modal RAG modeling, it is still hard to effectively learn more crucial semantics from these image features to improve the RAG performance within the multi-modal context that consists of retrieved documents.

\subsection{RAG Effectiveness within the Input Context of Different Modalities}
In this experiment, we investigate the impact of retrieved documents from different modalities on the effectiveness of RAG models.

As shown in Figure~\ref{fig:retrieval_modality_comparison}, we divide the multi-modal QA dataset of M$^2$RAG into two groups: image-answerable queries and text-answerable queries. Both categories represent queries that can be answered by image documents or text documents, respectively. We compare both vanilla RAG and MM-RAIT, implemented using MiniCPM-V and Qwen2-VL. Top5 ranked documents from texts, images, and both modalities are fed to different RAG models to evaluate their QA performance.
\begin{figure*}[t]
    \centering
    \subfigure[Image Captioning.\label{fig:appendix_case_image_cap}]{
        \includegraphics[width=0.48\linewidth]{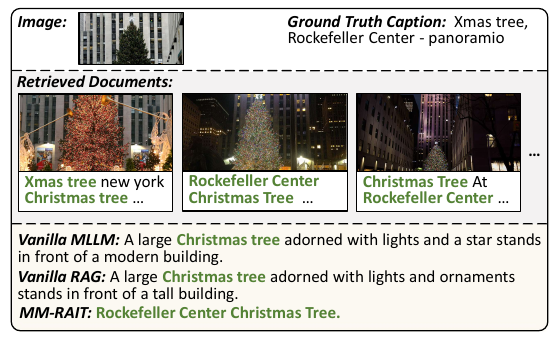}}
    \subfigure[Multi-Modal Question Answering.\label{fig:appendix_case_mm_qa}]{
        \includegraphics[width=0.48\linewidth]{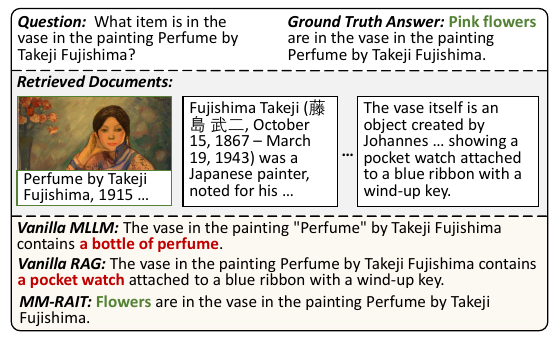}}
    \subfigure[Multi-Modal Fact Verification.\label{fig:appendix_case_mmfact_ck}]{
        \includegraphics[width=0.48\linewidth]{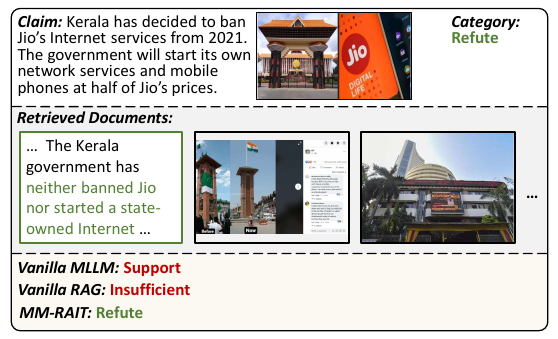}}
    \subfigure[Image Reranking.\label{fig:appendix_case_image_rerank}]{
        \includegraphics[width=0.48\linewidth]{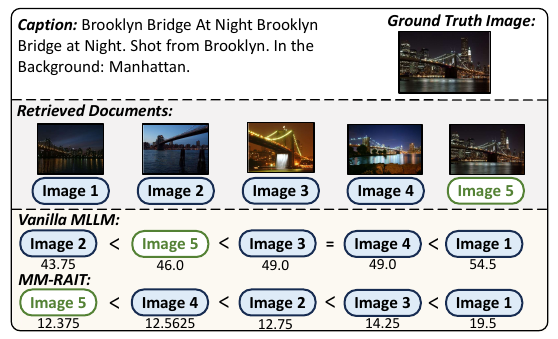}}
        
    \caption{Cases in Different Tasks. For generation tasks, we present the responses of different models using different RAG strategies (w/ or w/o RAG). We use green boxes to mark \textbf{\textcolor[HTML]{548235}{the documents that can provide information for the question}}. In the model output part, \textbf{\textcolor[HTML]{548235}{correct answers}} are marked in green, and red for \textbf{\textcolor[HTML]{C00000}{incorrect}}. For Image Reranking task, we presented the order reranked by different models through corresponding PPL scores.}
    \label{fig:case_different_k}
\end{figure*}

Figures~\ref{fig:text_qa_cpmv} and~\ref{fig:text_qa_qwen} show RAG performance on text-answerable queries. Overall, models using multi-modal retrieved documents perform similarly to those using only text, suggesting MLLMs can effectively learn from text sources. Vanilla RAG models show little variation across retrieval modalities, while MM-RAIT yields clear improvements with multi-modal inputs, demonstrating its ability to help MLLMs better leverage cross-modal context. Notably, vanilla MLLMs seem largely unaffected by retrieved content, likely relying more on internal knowledge for such queries.

Next, we evaluate the RAG performance on image-answerable queries, shown in Figures~\ref{fig:img_qa_cpmv} and~\ref{fig:img_qa_qwen}. The results indicate that RAG models using multi-modal documents generally outperform those using only text documents, confirming that incorporating image documents during retrieval enhances the ability of MLLMs to answer questions. The performance gap narrows for Qwen2-VL, suggesting that different MLLMs exhibit varying levels of reliance on multi-modal documents.

\subsection{Case Study}
As shown in Figure~\ref{fig:case_different_k}, in this section, we show four cases from Qwen2-VL in the four tasks of M$^2$RAG to evaluate the effectiveness of the MM-RAIT method within the multi-modal retrieval contexts. In the RAG setting, we use the top5 retrieved multi-modal documents for inference.

As illustrated in Figure~\ref{fig:appendix_case_image_cap}, in the image captioning task, both MLLM and vanilla RAG model tend to provide generic descriptions. After MM-RAIT training, Qwen2-VL extracts richer and more specific information from the retrieved multi-modal documents, such as ``Rockfeller Center'' landmark, generating more accurate captions. A similar improvement is observed in the image reranking task, as shown in Figure~\ref{fig:appendix_case_image_rerank}, where vanilla MLLMs initially struggle to align the semantics of the image and caption. After MM-RAIT training, fine-grained alignments between images and captions are achieved, allowing Qwen2-VL to rank the image of ``Brooklyn Bridge At Night. In the Background: Manhattan'' first, even though the reranking task is not involved during training.

For the multi-modal QA task shown in Figure~\ref{fig:appendix_case_mm_qa}, the question asks ``What item is in the vase in the painting Perfume by Takeji Fujishima?''
Due to the lack of background knowledge, vanilla MLLM generates an incorrect answer based on the query content, ``a bottle of perfume''. When multi-modal context is incorporated, the vanilla RAG model is influenced by irrelevant information in the last text document, ``a pocket watch'', leading to an incorrect answer. In contrast, MM-RAIT, benefiting from training, focuses more on the key document, extracts richer and more specific information, generating the correct answer ``Flowers''.
Similarly, in multi-modal fact verification, as shown in Figure~\ref{fig:appendix_case_mmfact_ck}, the vanilla RAG model struggles to extract useful information from noisy documents, while MM-RAIT enables the model to better extract and utilize relevant evidence, thereby improving its fact verification performance.

\section{Conclusion}
This paper proposes \textbf{M}ulti-\textbf{M}odal \textbf{R}etrieval-\textbf{A}ugmented \textbf{G}eneration (M$^2$RAG), a comprehensive benchmark designed to evaluate the capabilities of MLLMs in leveraging retrieved multi-modal contexts across four tasks. To further improve the effectiveness of retrieved information in generation, we also propose a \textbf{M}ulti-\textbf{M}odal \textbf{R}etrieval-\textbf{A}ugmented \textbf{I}nstruction \textbf{T}uning (MM-RAIT) method. MM-RAIT enhances MLLMs by explicitly optimizing them to process and integrate multi-modal retrieved content within an instruction-following framework, thereby improving their ability to utilize external multi-modal evidence during generation.


\begin{acks}
This work is partly supported by the National Natural Science Foundation of China (No. 62461146205), the Natural Science Foundation of China (No. 62206042), and the Fundamental Research Funds for the Central Universities (No. N25ZLL045). This work is also supported by the AI9Stars community.
\end{acks}
\clearpage

\bibliographystyle{ACM-Reference-Format}
\balance
\bibliography{sample-base}


\clearpage
\appendix

\twocolumn[{
\begin{center}
\includegraphics[width=0.95\textwidth]{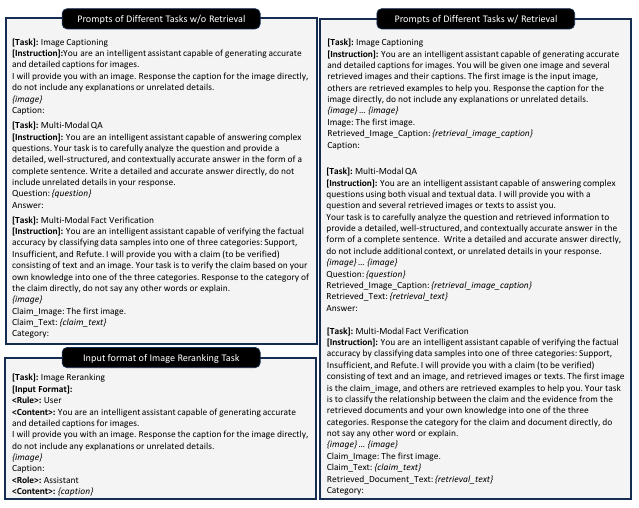}\captionof{figure}{Prompts Used for Different Tasks in Our M$^2$RAG Benchmark.}
\label{fig:prompts}
\end{center}
}]
\section{Appendix}
\subsection{License}
We show the licenses of the datasets that we use. WebQA uses CC0-1.0 license\footnote{\href{https://creativecommons.org/publicdomain/zero/1.0/}{https://creativecommons.org/publicdomain/zero/1.0/}}, while Factify uses MIT license\footnote{\href{https://opensource.org/licenses/MIT}{https://opensource.org/licenses/MIT}}.
All these licenses and agreements permit the use of their data for academic purposes.

\subsection{Prompt Templates Used in M$^2$RAG}\label{app:prompt} 
As shown in Figure~\ref{fig:prompts}, we present the prompt templates designed for various tasks in M$^2$RAG. Each task supports two settings: with and without retrieval, where retrieval refers to providing additional relevant images or text documents retrieved from the multi-modal corpus. In terms of image placement, we use the placeholder \emph{\{image\}} for the image, following the method proposed by~\citet{hu2024mrag}.

In the without-retrieval setting, prompts are concise and contain only the essential inputs—images, questions, or claims. For the image captioning task, the model directly generates a description based on the image. In multi-modal question answering, it responds solely based on the given question. For fact verification, the model determines the factuality of the claim using only its internal knowledge. While in the retrieval-augmented setting, prompts incorporate supplementary information such as retrieved images or text documents. They are designed to explicitly separate the primary input (e.g., the main image, question, or claim) from the retrieved evidence, guiding the model to leverage external context for more informed and accurate responses. 

We also define the input format for the image reranking task using a dialogue-style template, where the user provides the task prompt along with the retrieved image and the assistant generates the corresponding golden caption. This design enables MLLMs to effectively perform reranking image candidates by computing the Perplexity score.

\subsection{Additional Case Studies}
\begin{figure*}[t]
    \centering
    \includegraphics[width=1\linewidth]{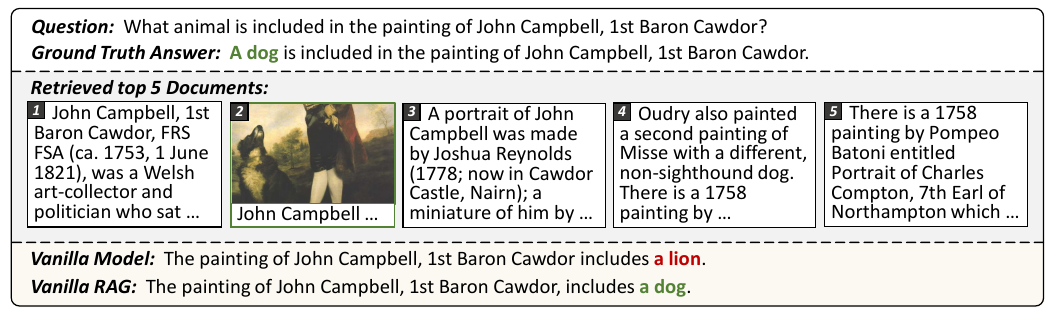} \\
    \includegraphics[width=1\linewidth]{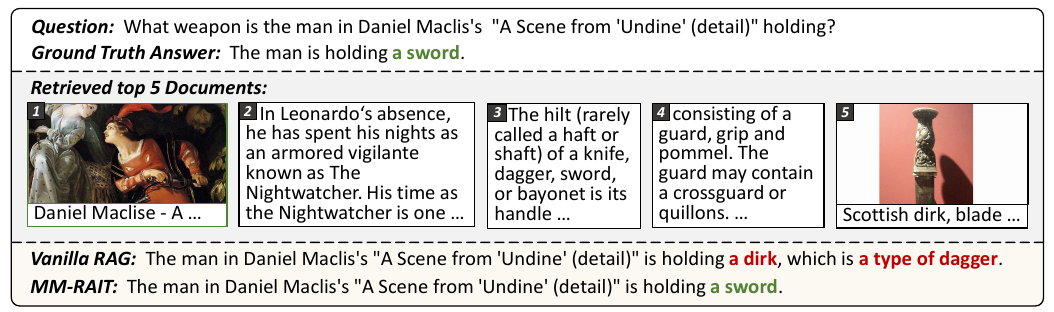}
    \caption{Additional Case Studies. We sample one example from each task from the M$^2$RAG benchmark and then show the performance of three models, including Vanilla MLLMs, Vanilla RAG models and MM-RAIT models. We also highlight the \textbf{\textcolor[HTML]{548235}{relevant phrases}}, \textbf{\textcolor[HTML]{548235}{correct answers}}, and \textbf{\textcolor[HTML]{C00000}{query-unrelated phrases}}.}
    \label{fig:case_additional_case}
\end{figure*}

In this section, we show two additional cases from Qwen2-VL in the Multi-Modal QA task of M$^2$RAG to evaluate the effectiveness of the MM-RAIT method within the multi-modal retrieval contexts.

As illustrated in Figure~\ref{fig:case_additional_case}, in the first case, the question asks, ``What animal is included in the painting of John Campbell, 1st Baron Cawdor?''. This requires the MLLM to match the ``1st Baron Cawdor'' and extract information about animals in the painting. Due to limited internal knowledge, the model encounters hallucination issues and generates an incorrect answer, ``a lion''. When the retrieved multi-modal document of ``1st Baron Cawdor'' is fed into the MLLM, the vanilla RAG model can directly extract ``dog'' from the painting, thus providing the correct response. This highlights the importance of multi-modal information in offering more intuitive and richer semantic insights to answer the question, underscoring the effectiveness of constructing the M$^2$RAG benchmark.

In the second case, the question asks that, ``What weapon is the man in Daniel Maclis's A Scene from `Undine' (detail) holding?'' Based on retrieved documents, the vanilla RAG model focuses on the fifth document, which depicts a ``Scottish dirk''. This leads the vanilla RAG model to generate an incorrect response, ``holding a dirk''. After MM-RAIT training, the model can accurately identify the relevant document describing the man holding a sword and extract pertinent information from it, thereby generating the correct response.

\subsection{Complete Evaluation Results of Additional MLLMs}
To enhance the representativeness of our benchmark and the MM-RAIT method, we provide supplementary evaluation results for both open-source MLLMs and API-based model on M$^2$RAG, as detailed in Table~\ref{tab:more_result}. For Open-source MLLMs, we include additional evaluations of Qwen2.5-VL 7B, InternVL-3 8B and LLaVA-NeXT-interleave-qwen 7B. For API-based MLLMs, GPT-4o-mini is assessed. Owing to recource limitations, when evaluating API-based models, 300 test instances are randomly sampled for each task. 
We also present the performance of MLLMs on the image reranking task, as shown in Table~\ref{tab:img_rerank}, where we tested the model performance under top3, top5 and top10 settings.
Furthermore, due to inherent constraints of API-based models, we can not obtain PPL values, precluding the acquisition of results for the image reranking task.

\begin{table*}[t]
\centering
\caption{\label{tab:more_result}Additional Overall Performance. We evaluate the performance of some mainstream MLLMs on our benchmark. Including Open-source MLLMs and API-based MLLMs.}
\resizebox{\textwidth}{!}{%
\begin{tabular}{l|ccc|ccc|cc}
\hline
\multirow{2}{*}{\textbf{Model}} & \multicolumn{3}{c|}{\textbf{Image Captioning}} & \multicolumn{3}{c|}{\textbf{Multi-Modal QA}} & \multicolumn{2}{c}{\textbf{MM Fact Verification}} \\ \cline{2-9} 
& \textbf{BERTScore} & \textbf{ROUGE-L}& \textbf{CIDEr}& \textbf{BERTScore} & \textbf{ROUGE-L}& \textbf{CIDEr}& \textbf{ ACC }& \textbf{F1} \\ \hline

\rowcolor{gray!8} \multicolumn{9}{l}{\textbf{Qwen2.5-VL (7B)}} \\ \hline
\textbf{Zero-Shot} & 85.89 & 19.94 & 22.07 & 78.18 & 37.69	& 139.96 & 47.70 & 39.64 \\ 
\textbf{Vanilla RAG} w/ top1 & 86.79 & 23.56 & 42.62 & 78.46	& 43.07	& 189.21 & 51.67 & 43.60 \\ 
\makebox[82pt][r]{w/ top3}& 86.86 & 23.92	& 42.88 & 78.39 & 43.53 & 190.85 & 53.73 & 47.08 \\ 
\makebox[82pt][r]{w/ top5} & 86.73 & 23.91 & 44.16 & 77.92 & 42.28 & 178.97 & 54.03 & 48.64 \\ 
\textbf{MM-RAIT} & \textbf{88.92} & \textbf{38.38} & \textbf{114.97} & \textbf{81.11} & \textbf{57.13} & \textbf{268.31} & \textbf{63.07} & \textbf{57.13} \\ \hline

\rowcolor{gray!8}\multicolumn{9}       {l}{\textbf{InternVL 3 (8B)}} \\ \hline
\textbf{Zero-Shot}& 85.45 & 17.15	& 18.87 & 78.02	& 35.51 & 106.09 & 45.47 & 45.08 \\ 
\textbf{Vanilla RAG} w/ top1& 87.15	& 27.83	& 69.91 & 78.30 & 42.48 & 176.62 & 55.90 & 56.45 \\ 
\makebox[82pt][r]{w/ top3}& 87.44 & 29.35	& 72.31 & 78.27	& 42.81 & 175.94 & 58.57 & 58.81  \\ 
\makebox[82pt][r]{w/ top5}& 87.23	& 28.52	& 65.65 & 78.36	& 43.26	& 182.04 & 59.67 & \textbf{59.86} \\ 
\textbf{MM-RAIT} & \textbf{88.69} & \textbf{36.07} & \textbf{101.13} & \textbf{81.10} & \textbf{64.09} & \textbf{333.92} & \textbf{62.30} & 58.26  \\ \hline

\rowcolor{gray!8} \multicolumn{9}{l}{\textbf{LLaVA-NeXT-interleave-qwen (7B)}} \\ \hline
\textbf{Zero-Shot} & 86.55	& 17.04 & 16.59 & 78.68 & 47.33	& 254.82 & 42.00 & 34.76  \\ 
\textbf{Vanilla RAG} w/ top1 & 87.14 & 22.51 & 34.96 & 78.75	& 49.58	& 265.24 & 52.03 & 43.14  \\ 
\makebox[82pt][r]{w/ top3}& 87.46 & 23.98	& 37.86 & 78.71 & 49.84 & 263.02 & 53.73 & 44.60  \\ 
\makebox[82pt][r]{w/ top5} & 87.48 & 24.03 & 37.14 & 78.69 & 49.82 & 263.71 & 53.60 & 44.35  \\ 
\textbf{MM-RAIT} & \textbf{88.49} & \textbf{35.04} & \textbf{97.85} & \textbf{80.71} & \textbf{58.93} & \textbf{298.31} & \textbf{62.83} & \textbf{57.27}  \\ \hline

\rowcolor{gray!8} \multicolumn{9}{l}{\textbf{GPT-4o-mini (randomly samples 300 instances per task)}} \\ \hline
\textbf{Zero-Shot} & \textbf{85.24}	& 17.11 & 15.36 & 77.92 & 33.92	& 85.37 & 53.67 & 49.20 \\ 
\textbf{Vanilla RAG} w/ top1 & 85.21 & 17.31 & 15.82 & \textbf{78.24}	& 39.90	& \textbf{140.59} & 56.33 & 56.73 \\ 
\makebox[82pt][r]{w/ top3}& 84.92 & \textbf{17.48}	& \textbf{17.31} & 78.11 & \textbf{40.40} & 129.88 & 60.67 & \textbf{60.03} \\ 
\makebox[82pt][r]{w/ top5} & 84.91 & 17.24 & 16.85 & 78.06 & 39.10 & 116.02 & \textbf{61.33} & 59.31 \\  \hline

\end{tabular}%
        }

\end{table*}

\begin{table}[t]
\caption{\label{tab:img_rerank}Image Reranking Task Result. We use FID↓ metric to evaluate the MLLMs performance.}
\begin{tabular}{l|ccc}
\hline
\multicolumn{1}{c|}{\multirow{2}{*}{\textbf{Model}}} & \multicolumn{3}{c}{\textbf{Image Reranking}}                           \\ \cline{2-4} 
\multicolumn{1}{c|}{}                       & \multicolumn{1}{c}{Top3} & \multicolumn{1}{c}{Top5} & Top10 \\ \hline
\textbf{Qwen2.5-VL (7B)}  & 10.51  & \textbf{10.29}  & 10.50      \\
MM-RAIT & 9.45 & 8.78  & \textbf{8.26}      \\ \hline
\textbf{InternVL 3 (8B)} & \textbf{11.08}  & 11.20   & 11.76      \\
MM-RAIT  & 10.76  & 9.49 & \textbf{9.36}       \\ \hline
\textbf{LLaVA-NeXT (7B)} & 10.52 & \textbf{10.51} & 10.67      \\
MM-RAIT & 10.55 & 9.65 & \textbf{9.43}      \\ \hline
\end{tabular}
\end{table}

\begin{table*}[t]
\centering
\caption{\label{tab:different_topk}Performance Under Different Topk Training Settings.}
\resizebox{\textwidth}{!}{%
\begin{tabular}{l|ccc|ccc|cc|c}
\hline
\multirow{2}{*}{\textbf{Model}} & \multicolumn{3}{c|}{\textbf{Image Captioning}} & \multicolumn{3}{c|}{\textbf{Multi-Modal QA}} & \multicolumn{2}{c|}{\textbf{MM Fact Verification}} & \textbf{Image Reranking} \\ \cline{2-10} 
& \textbf{BERTScore} & \textbf{ROUGE-L}& \textbf{CIDEr}& \textbf{BERTScore} & \textbf{ROUGE-L}& \textbf{CIDEr}& \textbf{ ACC }& \textbf{F1}& \textbf{FID↓} \\ \hline

\rowcolor{gray!8}\multicolumn{10}{l}{\textbf{MiniCPM-V 2.6 (8B)}} \\ \hline
\textbf{MM-RAIT}   w/ top1 & 88.37 & 35.26 & \textbf{99.67} & \textbf{81.08} & 57.64 & 292.68 & 61.17 & 56.56 & 9.46 \\ 
\makebox[72pt][r]{w/ top3}& 88.49 & \textbf{36.02} & 89.32 & 80.57 & 60.35 & 314.88 & \textbf{62.93} & \textbf{60.83} & 9.44 \\ 
\makebox[72pt][r]{w/ top5}& \textbf{88.50} & 35.61 & 89.18 & 80.52 & \textbf{61.32} & \textbf{318.39} & 61.60 & 58.47 & \textbf{9.28} \\ \hline

\rowcolor{gray!8}\multicolumn{10}{l}{\textbf{Qwen2-VL (7B)}} \\ \hline
\textbf{MM-RAIT}   w/ top1 & 88.89 & 38.06	& 111.40 & 80.84 & 59.09 & 310.44 & 62.27 & \textbf{58.09} & 8.76 \\ 
\makebox[72pt][r]{w/ top3}& 88.96 & 38.64	& 114.66 & \textbf{81.07}	& 61.76 & 325.24 & \textbf{63.00} & 57.60 & 8.69 \\ 
\makebox[72pt][r]{w/ top5}& \textbf{89.12} & \textbf{39.18}	& \textbf{118.00} & 80.46	& \textbf{63.45}	& \textbf{339.13} & 62.13 & 55.90 & \textbf{8.55} \\ \hline

\end{tabular}%
}

\end{table*}\label{app:ablation_topk}
\subsection{Ablation Study of Different TopK Training Setting}
To investigate the influence of diverse topK document specifications on the training efficacy of MMRAIT, we design analytical experiments under different topK configurations. To ensure consistency between the training and inference stages, models with varying topK settings will employ the corresponding topK strategies during the inference process. As for the image reranking task, models with different training strategies consistently rerank the top5 retrieved images.

As shown in Table~\ref{tab:different_topk}, In the tasks of multi-modal question answering, the performance of the model gradually improves as the topK value increases. This indicates that incorporating more contextual information during training requires the model to learn a stronger ability to filter effective information from the contexts. During inference, more contextual information can provide additional effective cues for generation. We also observed a subtle decrease trend in model performance on fact verification tasks as topK increases. We speculate that this is because the corpus for fact verification tasks is relatively small, and apart from the ground-truth document, there are no additional documents that can provide valid information. This leads to the introduction of extra invalid information into the model as topK increases when generation, thereby causing a decline in the model's performance on fact verification tasks.

\end{document}